\documentclass[a4paper,conference]{IEEEtran}

\usepackage{times}
\usepackage{graphicx}
\usepackage{latexsym}
\usepackage{subfig}
\usepackage{color}
\usepackage{amssymb}
\usepackage{multirow}
\usepackage{lipsum}

\ifCLASSINFOpdf
\else
\fi
\begin{document}
%
\title{Video-based Facial Expression Recognition using Graph Convolutional Networks}

\author{\IEEEauthorblockN{Daizong Liu}
\IEEEauthorblockA{Huazhong University \\ of Science and Technology \\
Email: dzliu@hust.edu.cn}
\and
\IEEEauthorblockN{Hongting Zhang}
\IEEEauthorblockA{Huazhong University \\ of Science and Technology \\
Email: htzhang@hust.edu.cn}
\and
\IEEEauthorblockN{Pan Zhou (Corresponding author)}
\IEEEauthorblockA{Huazhong University \\ of Science and Technology \\
Email: panzhou@hust.edu.cn}

\IEEEcompsocitemizethanks{
\IEEEcompsocthanksitem $^*$Co-First Author
}}

%


\maketitle

\begin{abstract}
Facial expression recognition (FER), aiming to classify the expression present in the facial image or video, has attracted a lot of research interests in the field of artificial intelligence and multimedia. In terms of video based FER task, it is sensible to capture the dynamic expression variation among the frames to recognize facial expression. However, existing methods directly utilize CNN-RNN or 3D CNN to extract the spatial-temporal features from different facial units, instead of concentrating on a certain region during expression variation capturing, which leads to limited performance in FER. In our paper, we introduce a Graph Convolutional Network (GCN) layer into a common CNN-RNN based model for video-based FER. First, the GCN layer is utilized to learn more significant facial expression features which concentrate on certain regions after sharing information between extracted CNN features of nodes. Then, a LSTM layer is applied to learn long-term dependencies among the GCN learned features to model the variation.
In addition, a weight assignment mechanism is also designed to weight the output of different nodes for final classification by characterizing the expression intensities in each frame. To the best of our knowledge, it is the first time to use GCN in FER task. We evaluate our method on three widely-used datasets, CK+, Oulu-CASIA and MMI, and also one challenging wild dataset AFEW8.0, and the experimental results demonstrate that our method has superior performance to existing methods.
\end{abstract}


%
\IEEEpeerreviewmaketitle

\section{Introduction}
\begin{figure}[t]
\centerline{\includegraphics[width=0.45\textwidth]{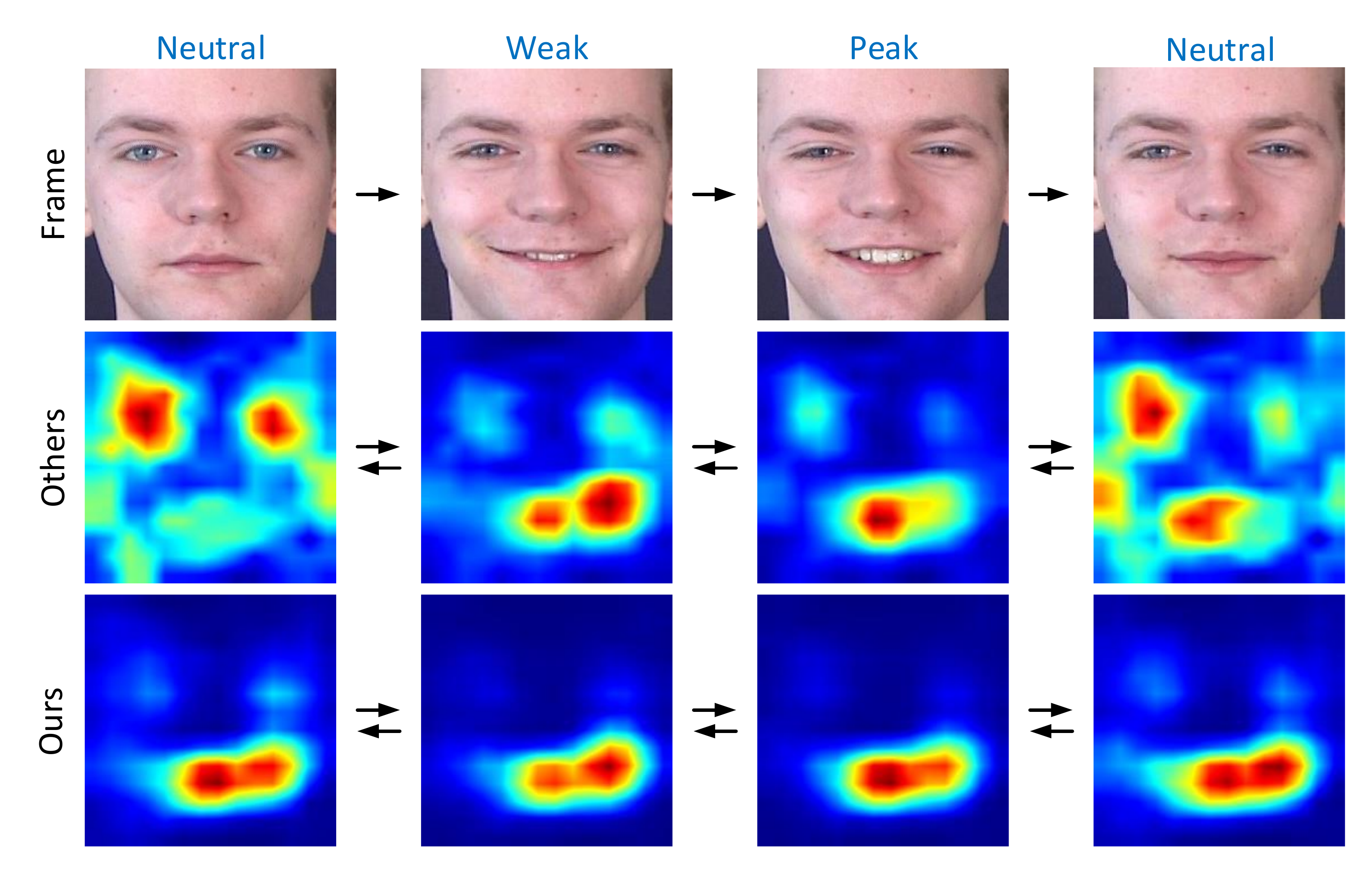}}
\caption{Video example of "Happy", where the expression starts from neutral stage to peak one and return to neutral again. The heatmap represents the concerned regions for expression recognition based on the learned features, where previous works (second row) focus on different regions in each frame of video while our method (bottom row) targets on a certain contributing expressional region for better variation exploring.} 
\label{fig:introduction}
\vspace{-12pt}
\end{figure}

Facial expression recognition (FER), as the task of classifying the emotion on images or video sequences \cite{bazzo2004recognizing,tong2007facial,ozbey2018expression,liu2016learning,monkaresi2016automated,zhang2017facial}, has become an increasingly dynamic topic in the field of computer vision in recent years. Although significant progress has been made towards improving the expression classification, there are still many challenges in exploring the dynamic expression variation. As shown in Fig. \ref{fig:introduction} (first row), the expression "Happy" is mostly contributed by the expressional intensity variation on the mouth region. Therefore, it is necessary to locate such informative region when capturing dynamic expression variation in video sequence.

Most of existing works \cite{bargal2016emotion,monkaresi2016automated,zhang2017facial} focus on extracting the feature representation of each frame using the Convolutional Neural Networks (CNN), which lacks a global consideration of correlation among all frames in video sequence. These methods aim to find out the most contributing expression features with each frame and take it as an image-based task by assembling these features to model the facial activation. 
Fig. \ref{fig:introduction} (second row) shows the individual features they learned from each frame, where different features focus on different regions. That is because the facial expression intensity on different regions is dynamically changing among the video frames. 
However, such features can only contribute limited strength to explore the dynamic variation of expression as they do not concentrate on the facial activation in an certain expression region (mouth).
Moreover, the features coming from peak frames usually focus on important regions which have more contributing information than those of non-peak frames. Therefore, there is a great need for guiding the mechanism to pay attention to the certain facial regions in all video frames, especially those focused by peak frames, to effectively capture the dynamic expression variation. 

Since Graph Convolutional Network (GCN) \cite{kipf2016semi,lee2018multi} has exhibited outstanding performances in learning correlative feature representations for specific tasks, it can be exploited to share the messages in graph and reconstruct the hidden states of each node to focus more on the significant information. We adapt GCN framework to FER task to learn the frame-based feature dependencies by training a learnable adjacency matrixs. After propagating expression features among the frame, GCN learn more contributing features due to the significant impact of peak frames on non-peak frames.

Although we learn expression features which focus on the same region in each frame to model the dynamic variation, those learned features of the peak frames still have more informative expressional representations than those of non-peak frames and should be considered more for final recognition.
To automatically distinguish peak frames in video-sequences, we characterize the expression intensities by deriving frame-wise weights from the elements of learned adjacency matrix in GCN layer.  
We utilize a weighted feature fusion function based on the expression intensity weights to integrate the reconstructed features. It can guide the model to focus on those peak expression frames which contribute more to the final classification.

To sum up, we propose a novel GCN based end-to-end framework for dynamic FER task, called Facial Expression Recognition GCN (FER-GCN), to learn more contributing facial expression features to capture dynamic expression variation. We introduce a GCN layer between CNN and RNN to achieve this. Firstly, our GCN layer updates the individual features of each frame based on the propagated features from the peak frames and learn an adjacency matrix which represents the inter-dependency among frames. 
With the GCN learned features focusing on the same regions, the LSTM layer is further applied to learn their long-term dependencies to model the variation. Fig. \ref{fig:introduction} (bottom row) shows GCN learned features which focus on the same region (mouth). 
Secondly, we adopt the learned adjacency matrix of GCN layer to represent expression intensities in time series. It can decrease the influence of the weak expressional features from neutral frames and exploit more expressional contributing ones from peak frames for final classification. Comparing to state-of-the-art approaches, our method is much more robust and achieves the best performances on four benchmarks (CK+, Oulu-CASIA, MMI and AFEW8.0).

Our main contributions are summarized as follows:
\begin{itemize}
    \item To the best of our knowledge, we are the first to apply GCN to FER task. Our graph based modules first propagate the most contributing expression features from peak frames among nodes to learn the frame-based features which focus on a certain expression region, and then explore the long-term dependencies among video frames to capture dynamic variation. It helps the model target on certain regions for expressional features learning.
    \item We also design a weighted feature fusion mechanism using adjacency matrix of GCN layer to fuse the features of all frames in one video sequence, where different learned weights represent different expression intensities of each frame, which eventually results in that the features of the peak frames contribute more to the final recognition while the weak expressional ones contribute less.
    \item We conduct our experiments on four public FER benchmark datasets, which demonstrates that the proposed method outperforms all state-of-the-art methods. And we also do ablation study which verified the effectiveness of each component in our model.
\end{itemize}

\section{Related Work}
Facial expression recognition (FER) has been studied over decades. Traditional researches \cite{bazzo2004recognizing,tong2007facial} either utilized facial fiducial points obtained by a Gabor-feature based facial point detector or focused on facial action units (AUs) directly \cite{tian2001recognizing,tong2007facial} to model temporal facial activations for FER task. As convolutional neural networks (CNN) can extract deeper and more contexual information, existing approaches which benefit from CNN can be generally divided into two categories: image-based and video-based.

\begin{figure*}
\centerline{\includegraphics[width=1.0\textwidth]{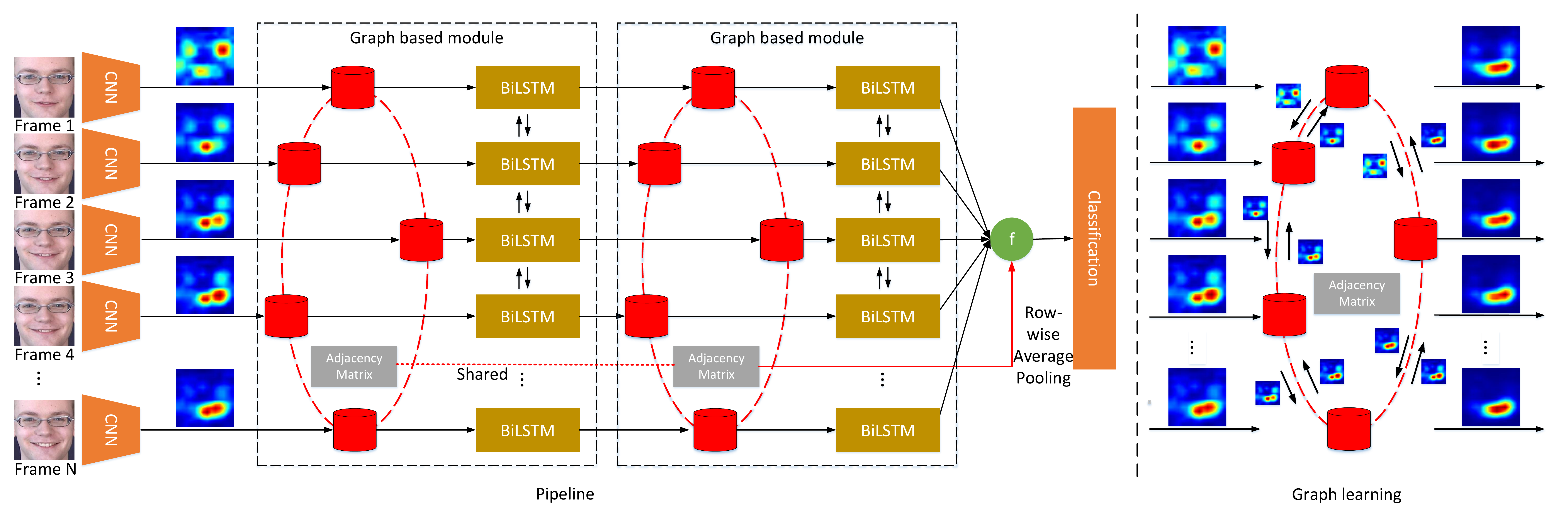}}
\caption{Overall architecture of our proposed method FER-GCN. Left:We apply two graph based modules to obtain the learned features by sharing the features among the frames, which focus on the most contributing regions for dynamic expression variation exploring. After that, we calculate the integrated representation for final classification by fusing them with the expression intensity weights learned from adjacency matrix $\textbf{\textit{A}}$ in graph. Right: details about how our GCN layer works. Each node shares its features to neighbors and updates itself with the matrix $\textbf{\textit{A}}$.} 
\label{fig:pipeline}
\vspace{-15pt}
\end{figure*}

Image-based methods \cite{ozbey2018expression,liu2016learning} do not consider dynamic variation and only study on still images. Yu \textit{et al.} \cite{yu2015image} proposed a method to exploit an ensemble multiple CNNs by minimizing a mixture of the log likelihood loss and the hinge loss. Bargal \textit{et al.} \cite{bargal2016emotion} established a hybrid network which combines VGG16 \cite{simonyan2014very} with residual neural network (RNN) to learn appearance features of expressions. Mollahosseini \textit{et al.} \cite{mollahosseini2016going} proposed to adopt three inception modules which have different critical considerations for a deeper and wider network. These image-based methods ignore the temporal information in a consecutive image sequence of facial expression, which plays an important role in capturing the dynamic variation for FER. To deal with this problem, a vast majority of works are explored toward video-based methods and have achieved remarkable performance. In video-based task \cite{monkaresi2016automated}, there is an additional capturing of dynamic variation of expression intensities among consecutive frames. Liu \textit{et al.} \cite{liu2014deeply} utilized 3D CNN to extract the spatio-temporal features and Zhang \textit{et al.} \cite{zhang2017facial} proposed a spatio-temporal network to extract dynamic-still information. Zhao \textit{et al.} \cite{zhao2016peak} also introduced that not all image frames in one video contribute equally to the final classification, and defined the peak and non-peak frames in the video sequences.

Although FER has shown good performance by video-based methods which successfully learn the temporal information among consecutive frames, it is still challenging when faced with the high intra-class variation. Some works introduced attention mechanism to their models to improve this situation. Minaee \textit{et al.} \cite{minaee2019deep} introduced attentional convolutional network into deep learning approach, which is able to focus on expressional parts of the face. Liu \textit{et al.} \cite{liu2019pose} proposed an attention mechanism in hierarchical scales to discover the most relevant regions to the facial expression, and select the most informative scales to learn the expression-discriminative representations. The introduction of attention module greatly improved the task performance over previous models on multiple datasets, but it is still not clear that how the expression features work or share in temporal domain in such module.

Inspired by works of Graph Convolutional Network (GCN) \cite{kipf2016semi,lee2018multi}, where each node shares the information with neighbors and then updates the state it learned based on the adjacency matrix, we develop a graph tailored to video-based FER task. Specifically, since our learnable adjacency matrix learned by the graph stands for how much each frame contributes to the final classification, we use it to distinguish peak frames from weak ones and reconstruct each node features during sharing the most contributing spatial expressive features to others. In the end, our method learns the most contributing spatial-temporal features in an interpretable way by graph learning, which leads to effective capture of the expressive component and proves to be more robust to individual variations.

\section{Methodology}
The architecture of our proposed method Facial Expression Recognition GCN (FER-GCN), illustrated in Fig. \ref{fig:pipeline} (left), is composed of four components: CNN based feature extraction module, graph based module, weighted features fusion module and the final classification. 
Given a facial video sequence $x_i$, $i={1,2,...,N}$ where $N$ is the number of frames, we first utilize a CNN network to extract their deep features. 
Then two graph based modules are following and each of them is exploited to learn more contributing expression features of each frames by a Graph Convolutional Network (GCN) layer and a Long Short Term Memory (LSTM) layer. 
At last, we derive $N$ weights of $N$ features from the learnable adjacency matrix of GCN layer, which implies the expression intensity of each frame, to fuse the $N$ features together for the final classification.

\subsection{Graph based Module}
To capture the dynamic expression variation more effectively, we propose a novel graph based module to capture the dynamic expression variation.
We build a GCN layer with $N$ frames, to propagate messages among the nodes in graph and model the frame-wise correlation by learning a dynamic adjacency matrix $\textbf{\textit{A}}$. All nodes tend to be influenced by expressional informative frames and update themselves as more contributing ones.
After the above process of graph learning, the $N$ frame updated features are further sent to the BiLSTM for long-term dependency learning in both forward and backward directions. The LSTM layer can capture the dynamic expression variation on certain concerned regions.

\textbf{Graph learning}
We first give the details about how our GCN layer works in Fig. \ref{fig:pipeline} (right). Our GCN layer contains $N$ nodes, which correspond to each frame of video sequence. 
During training GCN, we first generate the $N$ frame features $H_i \in \mathbb{R}^{1 \times d}$, $i={1,2,...,N}$ by CNN extractor or the previous GCN layer. 
Then we represent them as individual node to build a full-connected graph with a learnable adjacency matrix $\textbf{\textit{A}} \in \mathbb{R}^{N \times N}$. 
At every step, the GCN layer works in a way that each node shares its feature to neighbors and updates the state with both updated messages from neighbor nodes and the matrix $\textbf{\textit{A}}$ from the last time step. In fact, adjacency matrix $\textbf{\textit{A}}$ is dynamically updated with the backpropagated gradient in each time step, aiming to establish the inter-dependency among the frames.
The element $\textbf{\textit{A}}_{ij}$ in matrix $\textbf{\textit{A}}$ stands for how much the node $i$ depends on the node $j$, and thus the weak expression frames tend to have high possibility to depend on the peak ones for the latter focus on expressional region.
In this way, each node is more likely to update the features based on massages from the peak frame and thus focuses on the concerned expression region. The process of learning more contributing features can be formalized as the following.

For the $i$th node, it receives messages from the other $N-1$ neighbors, whose input features can be jointly represented as a matrix $\textit{n}_i \in \mathbb{R}^{(N-1)\times d}$ as follows:
\begin{equation}
    \textit{n}_i = [H_1^\mathrm{T} \\\ H_2^\mathrm{T} \\\ ... \\\ H_{i-1}^\mathrm{T} \\\ H_{i+1}^\mathrm{T} \\\ ... \\\ H_N^\mathrm{T}]^\mathrm{T}
\end{equation}
During the messages updating, the features from the neighbors are embedded with a learnable parameter matrix $\textbf{\textit{W}}^{l} \in \mathbb{R}^{d\times d}$ and then are propagated to node $i$. The embedded neighbors messages $M^{l}_i \in \mathbb{R}^{(N-1) \times d}$ can be calculated as follows:
\begin{equation}
    M^{l}_i = \textit{n}_i\textbf{\textit{W}}^{l}
\end{equation}
Here, $l$ represents the $l$th time step. Then the node $i$ updates its state by using both the updated messages $M^{l}_i$ and its own current state based on the $i$th row of the learned correlation matrix $\textbf{\textit{A}}$. Therefore, the output $o^{l+1}_i \in \mathbb{R}^{1 \times d}$ of node $i$ can be calculated as follows:
\begin{equation}
    \textbf{\textit{A}}_{i\bar{i}} = [\textbf{\textit{A}}_{i1},\textbf{\textit{A}}_{i2},\cdots,\textbf{\textit{A}}_{i(i-1)},\textbf{\textit{A}}_{i(i+1)},\cdots,\textbf{\textit{A}}_{in}]
\end{equation}
\begin{equation}
    o^{l+1}_i = f(\textbf{\textit{A}}_{i\bar{i}}M^{l}_i \oplus \textbf{\textit{A}}_{ii}H_i\textbf{\textit{W}}^{l})
\end{equation}
where $\textbf{\textit{A}}_{i\bar{i}} \in \mathbb{R}^{1\times(N-1)}$ is a matrix which consists of correlation coefficients between node $i$ and  the other nodes, and $\oplus$ means matrix addition. $f(\cdot)$ is the non-linear function like LeakyReLU. 
After updating the states of nodes into $o^{l+1} \in \mathbb{R}^{N \times d}$, where $d$ is the dimension of each node, the $N$ frame features are presented to focus on the same facial region as shown in Fig. \ref{fig:pipeline} (right), which indicates our GCN layer successfully guides the model to focus on the most contributing expression region among the video frames.

In addition, after the subsequent process of updating features, we get the loss and conduct the backpropagation. Our learnable adjacency matrix \textbf{\textit{A}} updates itself with the backpropagated gradient as follows:
\begin{equation}
    \textbf{\textit{A}}^{l+1} = \textbf{\textit{A}}^l - lr \ast \partial loss/\partial \textbf{\textit{A}}^l
\end{equation}
where $lr$ is the learning rate, and matrix \textbf{\textit{A}} will dynamically learn the inter-dependency among the frames to guide the message propagation in graph.

\textbf{Temporal variation modeling}
After processing the features by the GCN layer, the updated features in all frames focus on certain most contributing expression regions. 
Then, through the LSTM layer, we further learn the long-term temporal dependency for features concerned with certain regions in space. 
Specially, we adopt BiLSTM \cite{schuster1997bidirectional} to get access to the information from both past and future states for more contextual information combining.
Since the BiLSTM calculates the feature of each frame in each time step,
we give the output learned feature of each frame as follows:
\begin{equation}
    H^{l+1}_i = g(V_f\sigma (U_f[s^{l}_f, o^{l+1}_i]) + V_b\sigma(U_b[s^{l}_b, o^{l+1}_i]) + b), \\
    i \in [1,N]
\end{equation}
where $s_f^l,s_b^l \in \mathbb{R}^{d}$ are the hidden states containing information from previous and future time steps respectively. $U_f,U_b \in \mathbb{R}^{d/2 \times 2d}$ embed the concatenation of hidden state and input respectively in two directions. Then $V_f,V_b$ project embeddings from $\mathbb{R}^{d/2}$ to dimension $\mathbb{R}^{d}$. $b \in \mathbb{R}^{d}$ is the additional bias, and $g,\sigma$ are the activation functions $tanh$, $sigmoid$ respectively.

\textbf{Module details}
Note that, our GCN layer works by gathering messages from neighbor nodes based on the adjacency matrix $\textbf{\textit{A}}$, which is generally pre-defined in most researches. 
As matrix $\textbf{\textit{A}}$ is crucial for GCN training, we initial $\textbf{\textit{A}}$ with an identity matrix whose elements of main diagonal are 1 and the remaining are 0. It means that each frame is initialed to be independent at the beginning, and our graph will learn their dependencies during the graph updating.
And our LSTM layer learns the GCN output in $N$ steps respectively to explore the long-term dependency in time series.
Specially, we utilize two such graph based modules sharing the same adjacency matrix as a stacked structure for deep feature construction.

\subsection{Weighted Feature Fusion}
After passing two graph based modules, we get the learned features which are more informative than the initial CNN features owing to mainly focusing on the same regions on face.
However, there are still some learned features not informative enough, especially at the beginning of the video frames which usually has a weak expression. Therefore, we introduce a weight feature fusion mechanism to reemphasize the contribution of the peak ones.

\textbf{Expression intensity weights}
As the adjacency matrix \textbf{\textit{A}} learns the dependencies among the video frames, where the weak frames are more dependent on the peak frames, the relevant coefficients of the peak frames are larger than those of weak ones, which can represent the importance of individual frames among video based on their expression intensities. To represent the expression intensity of each frame, we develop a weight function based on the learned matrix \textbf{\textit{A}} to calculate corresponding frame-wise weights. Since the $i$th column of \textbf{\textit{A}} represents influence of the $i$th frame on other frames, the expression intensity weights can be formulated by :
\begin{equation}
    weight = softmax(mean(\textbf{\textit{A}}, dim=0))
\end{equation}
Here we apply row-wise average pooling on the matrix \textbf{\textit{A}} and a softmax function to get the normalized importance $weight \in \mathbb{R}^{1\times N}$ which represents the expression intensity in each frame.

\textbf{Fusion for final representation}
As the peak frames tend to contain more informative features than the weak ones, we need to reemphasize their different contributions for the final classification.
To focus more on the features of peak frame, we fuse the $N$ frame features $H_i, i=1,2,...,N$ with the expression intensity weight of each frame to generate the final representation. Our weighted feature fusion function and the final fused representation $r \in \mathbb{R}^d$ can be formulated as follows:
\vspace{-5pt}
\begin{equation}
    r = \sum_{i=1}^N weight_i H_i
\end{equation}
where the final representation $r$ can be calculated as the weighted sum of the feature sequence $H$ and the importance $weight$.

Note that since matrix \textbf{\textit{A}} not only participates in the graph learning, but also is utilized for the calculation of expression intensity weights. 
For correctly learning the graph correlation, we freeze the gradient of matrix \textbf{\textit{A}} in the weight calculation branch to avoid the gradient irrelevant to graph learning. 
We use values of the learned matrix \textbf{\textit{A}} to represent the intensities in dynamic expression variation. And we also clarify that the graph based module and weighted feature fusion are both indispensable to video-based FER task. The graph based module aims to learn the features based on the most contributing expression regions, which can guide the spatial module to focus on the most contributing expression region while some non-expressional features still exist in the weak frame. Thus our weighted feature fusion function helps to distinguish the peak and weak expression frames, to make the features of peak frame contribute more to the final recognition while decrease the impacts of the non-expressional features. Detailed visualization and analysis are illustrated in Section 4.4.

\section{Experiments}
In this section, we conduct the experiments on three widely used datasets, CK+ \cite{lucey2010extended}, Oulu-CASIA \cite{zhao2011facial}, and MMI \cite{pantic2005web}. We compare our model with state-of-the-art methods and do ablation study to demonstrate the effectiveness of each component in our model.
\subsection{Datasets}
Following the common evaluation strategy, we employ the most popular 10-fold cross-validation protocol on the following three datasets.

\textbf{CK+ dataset.} As an extended version of Cohn-Kanade (CK) dataset, this dataset includes 583 image sequences from 123 subjects, in which only 327  sequences from 118 subjects have facial expression labels (Anger, Contempt, Disgust, Fear, Happiness, Sadness and Surprise). For each of the video sequence, the intensity of the expression is reflected from neutral to the apex.

\textbf{Oulu-CASIA dataset.} It is composed of 6 basic facial expressions (Anger, Disgust, Fear, Happiness, Sadness and Surprise) from 80 subjects ranging from 23 to 58 years old. This dataset can be divided into 3 parts based on lighting conditions (normal, weak and dark), each of which consists of 480 sequences (80 subjects with 6 expressions). Similar
to CK+ dataset, all expression sequences begin at a neutral stage and end with the peak emotion.

\begin{table}
\centering
\caption{Average accuracy on the CK+, Oulu-CASIA and MMI datasets respectively.}
\label{tab:total_acc}
\begin{tabular}{ccccc}
\hline
\textbf{Method} & \textbf{CK+} & \textbf{Oulu} & \textbf{MMI} & \textbf{Feature} \\ \hline \hline
Inception \cite{mollahosseini2016going} & 93.20\% & - & 77.60\% & static \\ \hline
IACNN \cite{meng2017identity} & 95.37\% & - & 71.55\% & static \\ \hline
DLP-CNN \cite{li2017reliable} & 95.78\% & - & - & static \\ \hline
FN2EN \cite{ding2017facenet2expnet} & 96.80\% & 87.71\% & - & static \\ \hline
DeRL \cite{yang2018facial} & 97.30\% & 88.00\% & 73.23\% & static \\ \hline
PPDN \cite{zhao2016peak} & 99.30\% & 84.59\% & - & static \\ \hline
3DCNN \cite{liu2014deeply} & 85.90\% & - & 53.20\% & Dynamic \\ \hline
ITBN \cite{wang2013capturing} & 86.30\% & - & 59.70\% & Dynamic \\ \hline
HOG 3D \cite{klaser2008spatio}  & 91.44\% & 70.63\% & 60.89\% & Dynamic \\ \hline
TMS \cite{jain2011facial}  & 91.89\% & - & - & Dynamic \\ \hline
3DCNN-DAP \cite{liu2014deeply}  & 92.40\%  & - & 63.40\% & Dynamic \\ \hline
STM-ExpLet \cite{liu2014learning}  & 94.19\%  & 74.59\% & 75.12\% & Dynamic \\ \hline
LOMo \cite{sikka2016lomo} & 95.10\%  & 82.10\% & - & Dynamic \\ \hline
3D Inception-Resnet \cite{hasani2017facial} & 95.53\% & - & 79.26\% & Dynamic \\ \hline
Traj. on S+(2, n) \cite{kacem2017novel}  & 96.87\% & 83.13\% & 79.19\% & Dynamic \\ \hline
DTAGN \cite{jung2015joint}  & 97.25\% & 81.46\% & 70.24\% & Dynamic \\ \hline
GCNet \cite{kim2017deep} & 97.93\% & 86.11\% & 81.53\% & Dynamic \\ \hline
PHRNN-MSCNN \cite{zhang2017facial}  & 98.50\%  & 86.25\% & 81.18\% & Dynamic \\ \hline
\hline
\textbf{Ours} &  \textbf{99.54\%}  & \textbf{91.04\%} & \textbf{85.89\%} & Dynamic \\ \hline
\end{tabular}
\vspace{-15pt}
\end{table}

\textbf{MMI dataset.} This database includes 30 subjects of both genders and diverse ages from 19 to 62, containing 213 video sequences labeled with 6 basic expressions (Anger, Disgust, Fear, Happiness, Sadness, Surprise), out of which 205 sequences are with frontal face. And the expressions of subjects start from neutral state to the apex of one of the six basic facial expressions and return to the neutral state again.
\vspace{-5pt}

\subsection{Experimental Settings}
In our model, like most previous works, we set $N=16$ to choose $N$ frames chronologically from each video, and reuse frames if the number of whole frames less than 16. We utilize VGG16 \cite{simonyan2014very} with batch normalization layer as the feature extractor, which is initialized with the pre-trained model on ImageNet. In the graph based spatial-temporal module, we set the dimension $d$ of the feature vector in each node as 256, and we adopt LeakyReLU with the negative slope of 0.2 as the non-linear activation function followed by each GCN layer. We adopt BiLSTM \cite{schuster1997bidirectional} as the LSTM layer.

In the training phase, the input images are resized to $256 \times 256$ and then are randomly cropped into $224 \times 224$ with illumination changes and image flip for data augmentation. Our model is trained for 120 epochs with standard stochastic gradient descent (SGD) with learning rate set as 0.001 and weight decay set as 0.00005. We conduct all experiments using the Pytorch framework with a single NVIDIA 1080ti GPU.
\vspace{-5pt}

\subsection{Comparison to State-of-the-art Methods}
We use CK+ \cite{lucey2010extended}, Oulu-CASIA \cite{zhao2011facial}, and MMI \cite{pantic2005web} datasets for evaluation. We compare our method with state-of-the-art approaches which only use single end-to-end framework, not including the ensemble models like \cite{kuo2018compact,li2018deep}.

\begin{table}[t]
\centering
\caption{Confusion matrix of recognizing four expressions on CK+ dataset.}
\label{tab:ck_conf}
\begin{tabular}{c||ccccccc}
\hline
   & An & Co  & Di & Fe & Ha  & Sa  & Su  \\ \hline \hline
An & \textbf{100\%} & 0\% & 0\% & 0\% & 0\% & 0\% & 0\% \\ 
Co & 0\% & \textbf{100\%} & 0\% & 0\% & 0\% & 0\% & 0\% \\
Di & 0\% & 0\% & \textbf{100\%} & 0\% & 0\% & 0\% & 0\% \\
Fe & 0\% & 0\% & 0\% & \textbf{100\%} & 0\% & 0\% & 0\% \\
Ha & 0\% & 0\% & 0\% & 0\% & \textbf{100\%} & 0\% & 0\% \\
Sa & 0\% & 0\% & 0\% & 0\% & 0\% & \textbf{100\%} & 0\% \\
Su & 0\% & 1\% & 0\% & 0\% & 0\% & 0\% & \textbf{99\%} \\
\hline
\end{tabular}
\vspace{-8pt}
\end{table}

\begin{table}[t]
\centering
\caption{Confusion matrix of recognizing four expressions on Oulu-CASIA dataset.}
\label{tab:oulu_conf}
\begin{tabular}{c||cccccc}
\hline
   & An & Di & Fe & Ha  & Sa & Su \\ \hline \hline
An & \textbf{88\%} & 7\% & 1\% & 0\% & 3\% & 1\% \\ 
Di & 10\% & \textbf{84\%} & 2\% & 0\% & 3\% & 1\% \\
Fe & 0\% & 0\% & \textbf{91\%} & 4\% & 1\% & 4\% \\
Ha & 0\% & 0\% & 2\% & \textbf{98\%} & 0\% & 0\% \\
Sa & 4\% & 4\% & 1\% & 0\% & \textbf{90\%} & 1\% \\
Su & 0\% & 0\% & 4\% & 0\% & 1\% & \textbf{95\%} \\
\hline
\end{tabular}
\vspace{-10pt}
\end{table}

\begin{table}[t]
\centering
\caption{Confusion matrix of recognizing four expressions on MMI dataset.}
\label{tab:mmi_conf}
\begin{tabular}{c||cccccc}
\hline
   & An & Di & Fe & Ha & Sa & Su \\ \hline \hline
An & \textbf{77\%} & 13\% & 0\% & 0\% & 10\% & 0\% \\ 
Di & 3\% & \textbf{91\%} & 3\% & 0\% & 3\% & 0\% \\
Fe & 4\% & 0\% & \textbf{68\%} & 4\% & 4\% & 20\% \\
Ha & 0\% & 0\% & 2\% & \textbf{98\%} & 0\% & 0\% \\
Sa & 9\% & 0\% & 0\% & 0\% & \textbf{91\%} & 0\% \\
Su & 0\% & 0\% & 10\% & 0\% & 2\% & \textbf{88\%} \\
\hline
\end{tabular}
\vspace{-15pt}
\end{table}

\begin{figure*}
\centerline{\includegraphics[width=0.95\textwidth]{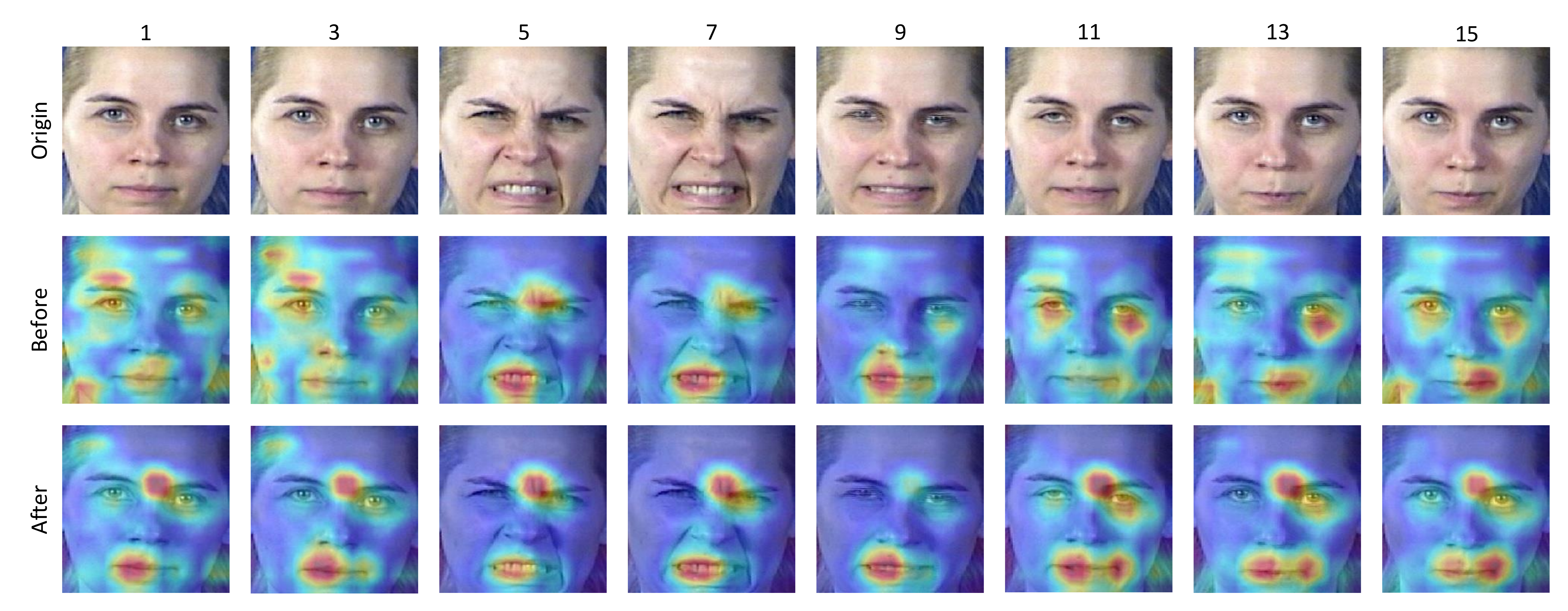}}
\caption{Example of the feature reconstruction in our GCN layer. First row: Origin facial images of "Disgust" in MMI dataset; Second row: input features of GCN layer; Third row: output features of GCN layer. It clarifies that our GCN layer shares most contributing expression features among frames to helps model focus more on the corresponding expression regions (such as mouth and nose here).} 
\label{fig:gcn}
\vspace{-15pt}
\end{figure*}

\textbf{Results on CK+}
As the results shown in Table \ref{tab:total_acc}, our proposed method takes the spatial-temporal feature propagation into consideration and achieves 99.54\% recognition rates on CK+ dataset, which outperforms the compared state-of-the-art methods in video task. Compared to PHRNN-MSCNN \cite{zhang2017facial}, which is also a video-based method, our model shows improvement of 1.04\%. Although PPDN \cite{zhao2016peak} treats video FER as the image-based task and only extracts the features from peak images to boost the performance of classification, it ignores noise of emotion changes in video sequences, and we outperform it by 0.24\%. The detailed confusion matrix on CK+ is given in Table \ref{tab:ck_conf}, where we find that almost all expressions are recognized well and "Surprise" shows the lowest recognition rate with 99\%.

\textbf{Results on Oulu-CASIA}
Compared to all the state-of-the-art methods on Oulu-CASIA dataset as shown in Table \ref{tab:total_acc}, our model achieves the best performance and has a 91.04\% accuracy rate. It outperforms PHRNN-MSCNN \cite{zhang2017facial} (video-based) and DeRL \cite{yang2018facial} (image-based) by 4.79\%, 3.04\% respectively. The confusion matrix in Table \ref{tab:oulu_conf} indicates that our method performs well in "Happiness" and "Surprise", but it shows the relatively low recognition rate with "Disgust", which is mostly confused with "Anger".

\textbf{Results on MMI}
Table \ref{tab:total_acc} also reports the comparison of our model with other state-of-the-art  methods on MMI dataset. Our model achieves the highest accuracy of 85.89\% and outperforms the previous best model GCNet \cite{kim2017deep} by 4.36\%. Compared to the PHRNN-MSCNN \cite{zhang2017facial}, which also utilizes the spatio-temporal representations, our method maps a expression variation graph to propagate the correlated features and has the improvement of 4.71\%. From the confusion matrix shown in Table \ref{tab:mmi_conf}, we can see that "Happiness" is relatively easy to be distinguished. "Anger" and "Fear" are mostly confused with "Disgust" and "Surprise", respectively.

\begin{figure*}[t]
\centerline{
\subfloat[CK+]{\includegraphics[width=0.32\textwidth]{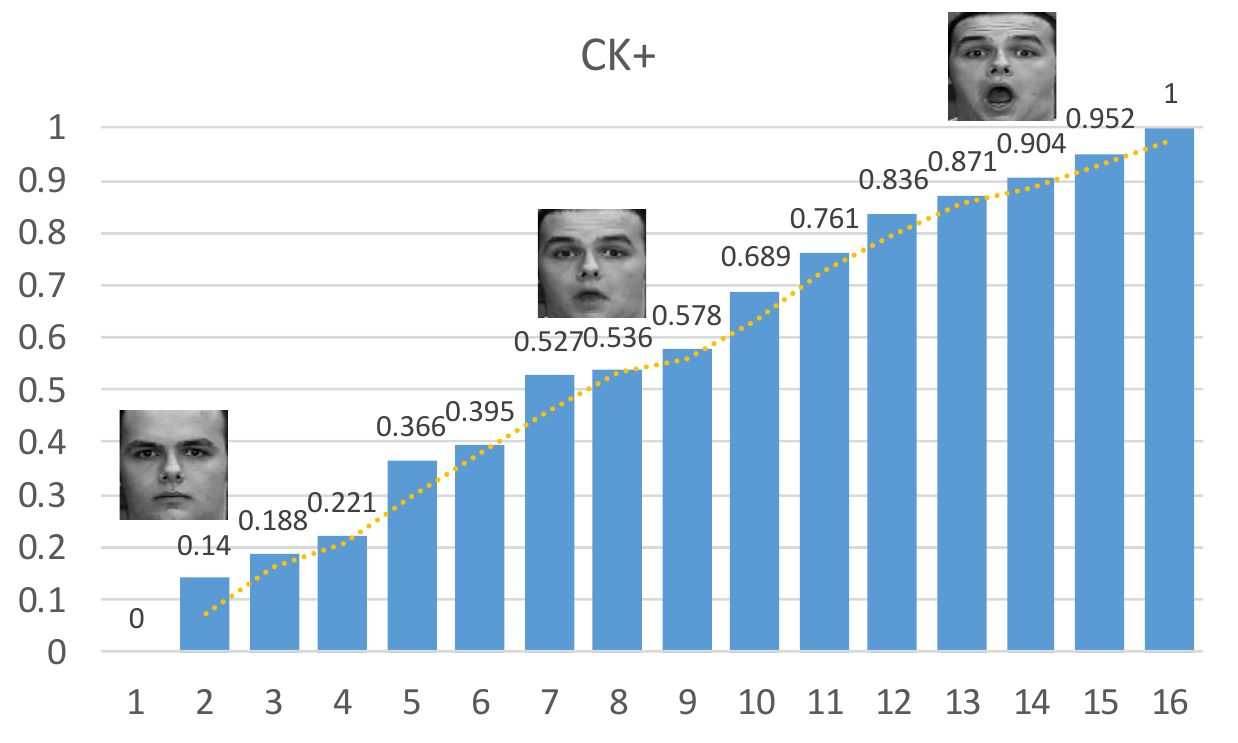}}
\subfloat[Oulu-CASIA]{\includegraphics[width=0.32\textwidth]{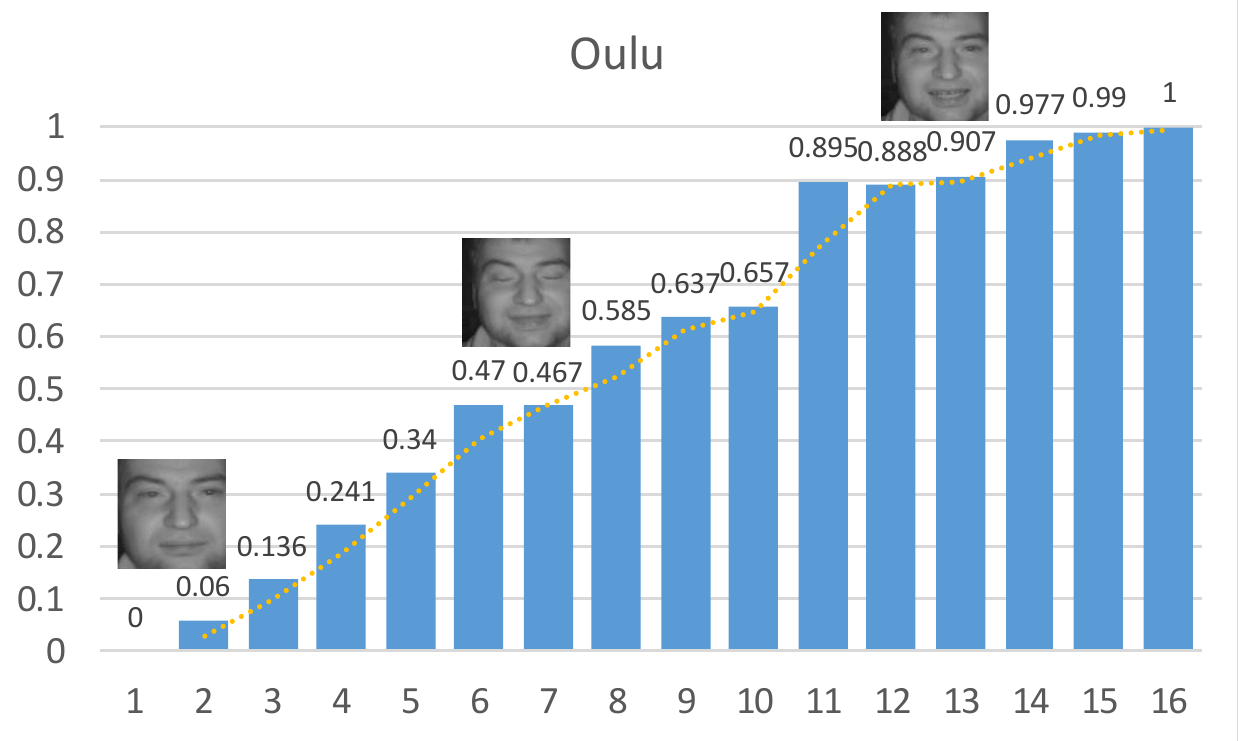}}
\subfloat[MMI]{\includegraphics[width=0.32\textwidth]{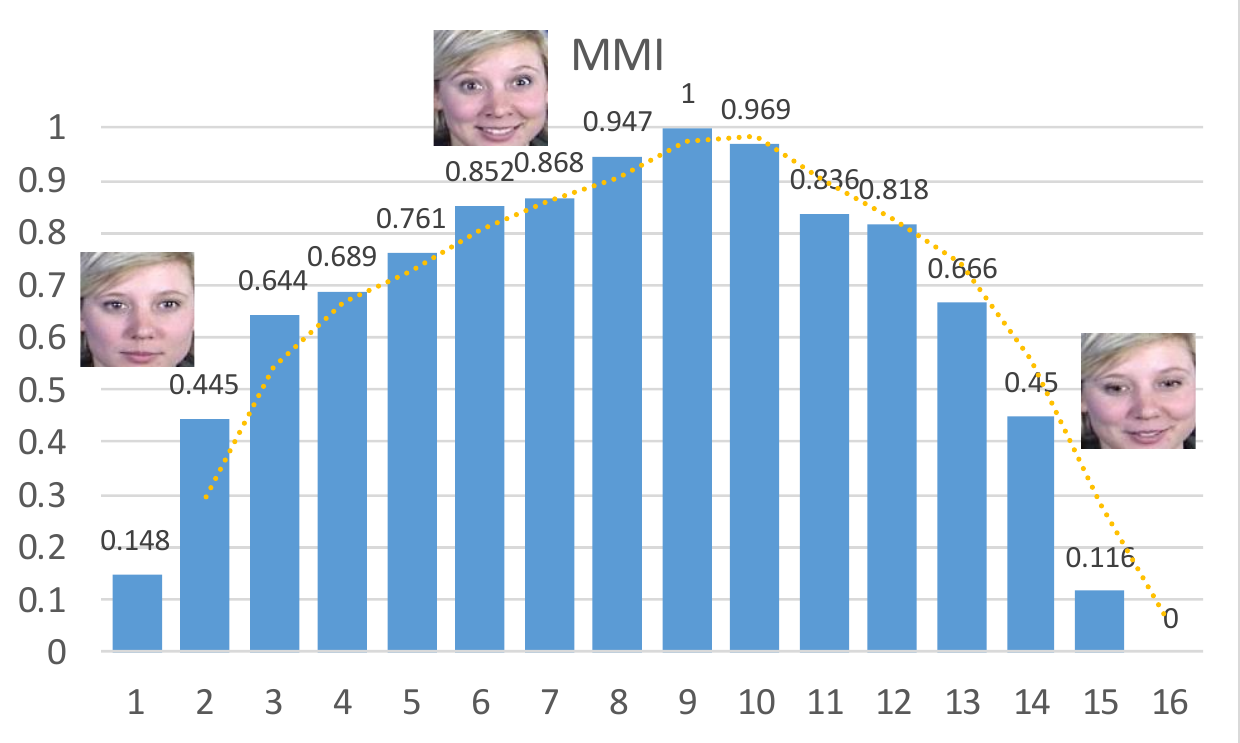}}}
\caption{Visualization of expression intensity weights for 16 steps on three datasets respectively. The horizontal axis represents the step number in each video sequence. The values of temporal weighs are given in the vertical axis through a sigmoid function, which refer to the expression intensity of each frame in the dynamic expression variation.}
\label{fig:weight}
\vspace{-12pt}
\end{figure*}

\subsection{Visualization and Analysis}
We further give the visualization to demonstrate the effectiveness of two components in our model: 1) we first show results of the GCN learned features which are updated with the propagated expression features in the graph based module; 2) and then we plot the expression intensity weights calculated from the learned adjacency matrix $\textbf{\textit{A}}$ in GCN layer to represent the expression intensity of each frame.

\textbf{GCN learned features}
In graph based module, we mainly illustrate how our GCN learns the $N$ frame features based on features from peak frames. As shown in Fig. \ref{fig:gcn}, the expression of origin facial images is "Disgust", whose expression intensity goes up from neutral to peak, then returns to neutral. The second row represents the extracted features from the previous CNN extractor, which shows that original CNN takes it as current image-based expression learning and concentrates on different facial parts in different frames. More in details, the weak frames (frame 1, 3, 11, 13, 15) focus on uncertain parts, while the peak frame (frame 5, 7, 9) mainly focus on the mouth and nose regions which are contributing more to the "Disgust" expression. We can see that, in the third row, features of all frames are learned to focus more on the mouth and nose regions with filtering out the non-expression contributing features. It demonstrates that our GCN layer shares the features among the video frames to guide them to pay attention to the most contributing expression region in all frames.

\textbf{Expression intensity weights}
The expression intensity weights represent the expression intensity of each frame among a video sequence, where the weights of peak frames tend to be larger and the weak ones smaller. We give the visualization of the expression intensity weights learned by adjacency matrix $\textbf{\textit{A}}$ in GCN layer on three datasets in Fig. \ref{fig:weight} respectively, where we normalize the weights through a sigmoid function for better understanding. We find that the weights of CK+ and Oulu-CASIA increase gradually from the first frame to the last frame in video sequence while the weights of MMI achieve highest value in the middle part. It demonstrates that our adjacency matrix $\textbf{\textit{A}}$ which relies on expression intensities among the dynamic expression variation, is able to learn the dependencies between frames and can help our model to automatically locate the peak expression frames in video FER task.
\vspace{-3pt}

\subsection{Ablation Study}
We run an extensive ablation study to demonstrate the effectiveness of different components of our proposed model FER-GCN, including the components of graph based spatial-temporal module and weighted feature fusion function. 

\begin{table}
\centering
\caption{Ablation study on the individual components.}
\label{tab:ablation1}
\begin{tabular}{cccc}
\hline
Experiment model & CK+ & Oulu-CASIA & MMI \\ \hline \hline
\multirow{2}*{VGG16} & \multirow{2}*{97.78\%} & \multirow{2}*{85.83\%} & \multirow{2}*{80.75\%} \\ 
~ & ~ & ~ & ~ \\ \hline 
VGG16 + graph based & \multirow{2}*{98.39\%} & \multirow{2}*{88.33\%} & \multirow{2}*{84.37\%} \\
spatial-temporal module$\times1$ & ~ & ~ & ~ \\ \hline
VGG16 + graph based & \multirow{2}*{99.09\%} & \multirow{2}*{89.79\%} & \multirow{2}*{84.64\%} \\
spatial-temporal module$\times2$ & ~ & ~ & ~ \\ \hline
VGG16 + graph based & \multirow{2}*{99.00\%} & \multirow{2}*{87.71\%} & \multirow{2}*{83.07\%} \\
spatial-temporal module$\times3$ & ~ & ~ & \\ \hline
VGG16 + graph based & \multirow{3}*{\textbf{99.54\%}} & \multirow{3}*{\textbf{91.04\%}} & \multirow{3}*{\textbf{85.89\%}} \\
spatial-temporal module$\times2$ & ~ & ~ & ~ \\
+ weighted feature fusion & ~ & ~ & ~ \\ \hline
\end{tabular}
\vspace{-15pt}
\end{table}

\textbf{Ablation study on individual components} We first give the study on the contributions of individual components in our model As shown in Table \ref{tab:ablation1}, the VGG16 backbone achieves the accuracy of 97.78\%, 85.83\% and 80.75\% on three datasets, which outperforms some existing methods because of our designed training process. With the spatial-temporal feature propagation and reconstruction, the \textit{VGG16+graph based spatial-temporal module$\times1$} outperforms the backbone by 0.61\%, 2.50\% and 3.62\% on three datasets respectively. It demonstrates that the graph based module helps to guide our model to focus on the peak expression regions among video frames to explore the dynamic expression variation for final recognition. Also, we find that the performance of FER achieves the highest accuracy of 99.09\%, 89.79\% and 84.64\% with only two graph based spatial-temporal modules and it is not going better when we utilize more. We give the analysis that the propagation between the nodes will be accumulated if we use more GCN layers, and it will result in over-smoothing. That is, the node features may be over-smoothed such that the features of nodes with different expression intensities may become indistinguishable. At last, our weighted feature fusion function has another improvement of 0.45\%, 1.25\% and 1.25\% on three datasets respectively, which shows its strong ability to capture the dynamic expression variation in video sequence.

\vspace{-5pt}
\subsection{Additional Evaluation on Wild Database}
At last, we conduct an additional experiment on a public "in the wild" dataset AFEW 8.0 \cite{dhall2018emotiw} to further investigate the robustness of our proposed method. In details, we follow the data pre-processing by \cite{lu2018multiple} and only compare our FER-GCN with the top-ranked single models or baselines in Emotiw2018 \cite{dhall2018emotiw} on the validation set. As shown in Table \ref{tab:wild}, the baseline of Emotiw2018 achieves the lowest performance of 38.81\% where the other methods have large improvement with deep feature extractor and temporal feature exploring. Although VGG-Face-LSTM achieves the performance of 53.91\% by exploiting spatial-temporal features, our proposed FER-GCN explores more interpretable features from the most contributing expression regions among the frames to capture the dynamic variation, and outperforms it by 1.76\%. It indicates that our proposed model helps to learn a more general dynamic expressional feature representation.
\begin{table}
\centering
\caption{Recognition accuracy of each single model on the validation dataset of AFEW 8.0.}
\label{tab:wild}
\begin{tabular}{|c|c|}
\hline
Method & Accuracy \\ \hline
Emotiw2018 (baseline) \cite{dhall2018emotiw} & 38.81\% \\
HoloNet \cite{hu2017learning} & 46.50\% \\
DSN-VGG-Face \cite{fan2018video} & 48.04\% \\
Resne50-LSTM \cite{lu2018multiple} & 49.31\% \\
DenseNet161-pool5 \cite{liu2018multi} & 51.44\% \\
VGG-Face-LSTM \cite{lu2018multiple} & 53.91\% \\ \hline
Ours & \textbf{55.67\%} \\ \hline
\end{tabular}
\vspace{-15pt}
\end{table}
\section{Conclusion}
In this paper, we present a novel framework named FER-GCN, which utilizes graph work to learn most contributing features for facial expression recognition. Our designed graph based module learn features of each node based on the propagated features from peak frames for long-term dependency exploring. And the adjacency matrix learned from the GCN layer is further applied to locate the peak frame in video sequence and further guide our model to focus on features of the peak frame. Experimental results on four widely used facial expression datasets demonstrate the superiority of our method compared with other state-of-the-art methods. 

\bibliographystyle{IEEEtran}

\bibliography{IEEEexample}

\end{document}